\documentclass{article}
\usepackage{iclr2021_conference,times}

\usepackage{amsmath,amsfonts,bm}

\def\eqref#1{equation~\ref{#1}}

\def\1{\bm{1}}

\DeclareMathAlphabet{\mathsfit}{\encodingdefault}{\sfdefault}{m}{sl}
\SetMathAlphabet{\mathsfit}{bold}{\encodingdefault}{\sfdefault}{bx}{n}

\usepackage{hyperref}
\usepackage{url}
\usepackage{graphicx}
\usepackage{multirow}

\title{Fair Interpretable Learning via Correction Vectors}

\newcommand*\samethanks[1][\value{footnote}]{\footnotemark[#1]}

\author{Mattia Cerrato\thanks{These authors contributed equally.}, Marius Köppel\samethanks[1], Alexander Segner\samethanks[1] ~\& Stefan Kramer \\
Johannes Gutenberg-Universität Mainz\\
Saarstraße 21, Mainz, Germany \\
}

\iclrfinalcopy

\begin{document}

\maketitle

\begin{abstract}
Neural network architectures have been extensively employed in the fair representation learning setting, where the objective is to learn a new representation for a given vector which is independent of sensitive information. Various ``representation debiasing'' techniques have been proposed in the literature. However, as neural networks are inherently opaque, these methods are hard to comprehend, which limits their usefulness. We propose a new framework for fair representation learning which is centered around the learning of ``correction vectors'', which have the same dimensionality as the given data vectors. The corrections are then simply summed up to the original features, and can therefore be analyzed as an explicit penalty or bonus to each feature. We show experimentally that a fair representation learning problem constrained in such a way does not impact performance.
\end{abstract}

\section{Introduction}

The issue of fairness in machine learning relates to analyzing the outcomes of automated decision systems which may impact people's well-being. In group fairness, one is dealing with statistically disparate outcomes for individuals belonging to different groups (e.g., women and men, black and white people) (\cite{zafar2017fairness}). One case which attracted much attention is the COMPAS software for recidivity prediction, which can be seen as biased against black people\footnote{For a complete view of the Northpointe/ProPublica debate, we refer the reader to the original report by  \cite{machine_bias} and Northpointe's rebuttal by \cite{dieterich2016compas}.}. Focusing on the fact that biased models derive from biased data, many authors have focused on learning fair representations for individuals (\cite{zemel2013learning, cerrato2020constraining, moyer2018invariant, chouldechova2017fair, fair_autoencoder}). In this setting, a representation algorithm such as a feedforward neural net trained via backpropagation is paired with an explicit fairness objective. Previous proposals have employed Maximum Mean Discrepancy (\cite{fair_autoencoder, gretton2012kernel}), adversarial learning (\cite{cerrato2020constraining, ganin2016jmlr, xie2017controllable}) and Mutual Information bounds (\cite{moyer2018invariant}). 

One issue with fair representation learning algorithms based on neural networks is their opaqueness. These methods project the original data space $X$ into a latent space $Z$ whose dimensions are incomprehensible to humans, as they are non-linear combinations of the original features. This is a noteworthy problem especially in the context of the ``right to an explanation'' as required in the EU by the GDPR, Recital 71. Therefore, these methodologies might be inapplicable in the real world.

In this context, we propose a new fair representation learning framework which learns \emph{feature corrections} instead of an entirely new space of opaque parameters.
In practice, this is akin to a pre-processing technique which changes the original features so to balance them between individuals belonging to different groups.
This guarantees a ``right to an explanation'' in the sense that it is always possible to extract the ``fair correction'' that has been applied to the data belonging to each individual. Furthermore, as the correction is computed via neural networks, our framework still enjoys all the benefits of the universal approximation theorems (see \cite{cybenko1989approximation}) and may therefore compute any debiasing function. 
Our framework is flexible, as it imposes only architectural constraints on the neural network without impacting the training objective: therefore, all neural debiasing methodologies may be extended so to belong in our framework.

Our contributions can be summarized as follows:

\begin{itemize}
    \item We develop a new family of fair representation learning algorithms based on neural networks.
    Our framework is interpretable as it relies on computing ``correction vectors'', which are simply added to the original representation.
    \item We discuss how to modify various existing algorithms so that they may belong in our framework.
    \item We show that extending a state-of-the-art fair representation learning algorithm to be interpretable does not affect performance negatively on both relevance and debiasing.
\end{itemize}

\section{The Interpretable Fair Framework}

In this section we describe our framework for interpretable fair representation learning.
Our framework makes interpretability possible by means of computing~\emph{correction vectors}.
Commonly, the learning of fair representations is achieved by learning a new feature space $Z$ starting from the input space $X$.
To this end, a parameterized function $f_\theta(x)$ is trained on the data and some debiasing component which looks at the sensitive data $s$ is included.
After training, debiased data is available by simply applying the learned function $z = f_\theta(x)$.
Any off-the-shelf model can then be employed on the debiased vectors.
Various authors have investigated techniques based on different base algorithms.

The issue with the aforementioned strategy is one of interpretability.
While it is possible to guarantee \emph{invariance to the sensitive attribute} -- with much effort -- by training classifiers on the debiased data to predict the sensitive attribute, it is unknown what each of the dimensions of $Z$ represent.
Depending on the relevant legislation, this can severely limit the applicability of fair representation learning techniques in industry.
Our proposal is to mitigate this issue by instead \emph{learning fair corrections} for each of the dimensions in $X$.
Fair corrections are then added to the original vectors so that the semantics of the algorithm are as clear as possible.
For each feature, an individual will have a clear penality or bonus depending on the sign of the correction.
Thus, we propose to learn the latent feature space $Z$ by learning fair corrections $w$: $w = f_\theta(x)$ and $z = w + x$. 

\begin{figure}
    \centering
    \includegraphics[scale=0.5]{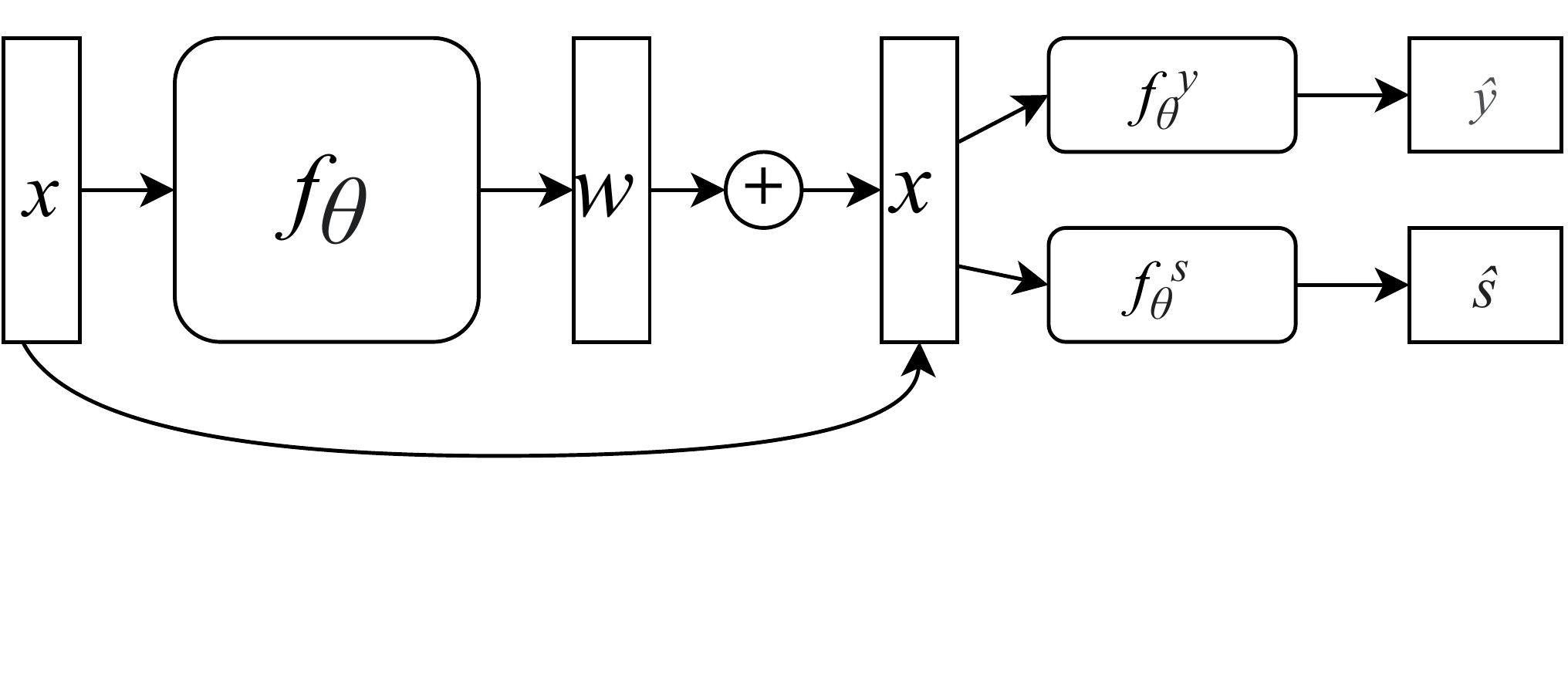}
    \caption{A gradient reversal-based neural network constrained for interpretability so to belong in our interpretable framework. The vector $w$ matches in size with $x$, and can then be summed with the original representation $x$ and analyzed for interpretability. This architectural constraint can be applied to other neural architectures.}
    \label{fig:architecture}
\end{figure}

It is very practical to modify existing neural network architectures so that they can belong in the aforementioned framework.
While there are some architectural constraints that have to be enforced, the learning objectives and training algorithms may be left unchanged.
The main restriction is that only ``autoencoder-shaped'' architectures may belong in our framework.
Plainly put, the depth of the network is still a free parameter, just as the number of neurons in each hidden layer. However, to make interpretability possible, the last layer in the network must have the same number of neurons as there are features in the dataset.
In a regular autoencoder architecture, this makes it possible to train the network with a ``reconstruction loss'' which aims for the minimization of the difference between the original input $x$ and the output $\hat{x} = f(x)$, where $f$ is a neural network architecture.
This is not necessarily the case in our framework.
On top of this restriction, we also add a parameter-less ``sum layer'' which adds the output of the network to its input, the original features.
Another way to think about the required architecture under our framework is as a skip-connection in the fashion of ResNets (\cite{he2016deep}) between the input and the reconstruction layer (see Figure~\ref{fig:architecture}).

Constraining the architecture in the aforementioned way has the effect of making it possible to interpret the neural activations of the last layer in feature space. 
As mentioned above, our framework is flexible in the sense that many representation learning algorithms can be constrained so to enjoy interpretability properties.
To provide a running example, we start from the debiasing models based on the Gradient Reversal Layer of ~\cite{ganin2016jmlr} originally introduced in the domain adaptation context and then employed in fairness by various authors (e.g. \cite{ mcnamara2017provably, xie2017controllable}).
The debiasing effect here is enforced by training a subnetwork $f^s_{\theta_1}(z)$ to predict the sensitive attribute and inverting its gradient when backpropagating it through the main network $f(x)$.
Another sub-network learns to predict $\hat{y} = f^y_{\theta_2}(z)$.
Both networks are connected to a main ``feature extractor'' $z = f_{\theta_3}(x)$.
The two models are pitted against one another in extracting useful information for utility purposes (estimating $p(y \mid \hat{x})$) and removing information about $s$ (which can be understood as minimizing $I(\hat{x}, s)$, see \cite{cerrato2020constraining}).
Here no modification is needed to the learning algorithm, while the architecture has to be restricted so that the length of the $\hat{x}$ vector is the same as the original features $x$.
One concerning factor is whether the neural activations can really be interpreted in feature space, as features can take arbitrary values or be non-continuous (e.g. categorical).
We circumvent this issue by coupling the commonly employed feature normalization step and the activation functions of the last neural layer.
More specifically, the two functions must map to two coherent intervals of values.
As an example, employing standard scaling (feature mean is normalized to 0, standard deviation is normalized to 1) will require an hyperbolic tangent activation function in the last layer.
The model will then be enabled in learning a negative or positive correction depending on the sign of the neural activation.
It is still possible to use sigmoid activations when the features are normalized in $[0, 1]$ by means of a min-max normalization (lowest value for the feature is 0 and highest is 1).
Summing up, the debiasing architecture of Ganin et al. can be modified via the following steps:

\begin{enumerate}
    \item Normalize the original input features $x_{raw}$ via some normalization function $x = g(x_{raw})$.
    \item Set up the neural architecture so that the length of $w = f_{\theta_3}(x)$ is equal to the length of $x$.
    \item Add a skip-connection between the input and the reconstruction layer.
\end{enumerate}

After training, the corrected vectors $z = f_{\theta_3}(x) + x$ and the correction vectors $w = f_{\theta_3}(x)$ can be interpreted in feature space by computing the inverse normalization $\hat{x}_{raw} = g^{-1}(z)$ and $w_{raw} = g^{-1}(w)$.  

Other neural algorithms can be modified similarly so to belong in the interpretable fair framework, and similar steps can be applied to e.g. the Variational Fair Autoencoder by \cite{fair_autoencoder} and the variational bound-based objective of \cite{moyer2018invariant}. In our experiments, we will however focus on the  state-of-the-art fair ranking model of \cite{cerrato2020pairwise}, which is based on the gradient reversal layer by \cite{ganin2016jmlr}.

\section{Experiments}

In the experiments we constrain the fair ranking model of \cite{cerrato2020constraining} to belong in our framework. This ranker employs the gradient reversal concept introduced in~\cite{ganin2016jmlr}. Therefore, as explained in depth in Section 3, it is sufficient to constrain its architecture to extract features which have the same dimensionality as $X$. After adding the skip-connection between the first layer and the last feature extraction layer, no other changes are needed to the training algorithm, which we leave unchanged from the original work. Therefore, we train the model employing SGD and select hyperparameters (the number of hidden layers; the fairness-relevance parameter $\gamma$; the learning rate for SGD) employing a nested cross-validation with $k=3$. We relied on the Bayesian Optimization implementation provided by Weights \& Biases (\cite{wandb}) and stopped after 200 model fits. To evaluate the models, we computed their nDCG, rND (a disparate impact fairness metric defined in~\cite{ke2017measuring}) and GPA (a disparate mistreatment metric defined in~\cite{fair_pair_metric}) in Table 1. We selected the best models with the assumption that each metric has equal importance.

\begin{table}[]
\centering
\begin{tabular}{cllllll}
\cline{3-4}
\multicolumn{1}{l}{}                          & \multicolumn{1}{l|}{}         & \multicolumn{1}{l|}{Fair DR}              & \multicolumn{1}{l|}{Interpretable Fair DR} &  &  &  \\ \cline{1-4}
\multicolumn{1}{|c|}{\multirow{3}{*}{COMPAS}} & \multicolumn{1}{l|}{1-rND}    & \multicolumn{1}{l|}{0.841411 $\pm$ 0.073} & \multicolumn{1}{l|}{0.822243 $\pm$ 0.065}  &  &  &  \\ \cline{2-4}
\multicolumn{1}{|c|}{}                        & \multicolumn{1}{l|}{1-GPA}    & \multicolumn{1}{l|}{0.927383 $\pm$ 0.034} & \multicolumn{1}{l|}{0.939985 $\pm$ 0.036}  &  &  &  \\ \cline{2-4}
\multicolumn{1}{|c|}{}                        & \multicolumn{1}{l|}{nDCG@500} & \multicolumn{1}{l|}{0.474789 $\pm$ 0.085} & \multicolumn{1}{l|}{0.526513 $\pm$ 0.067}  &  &  &  \\ \cline{1-4}
\multicolumn{1}{|c|}{\multirow{3}{*}{Bank}}   & \multicolumn{1}{l|}{1-rND}    & \multicolumn{1}{l|}{0.813426 $\pm$ 0.004} & \multicolumn{1}{l|}{0.812811 $\pm$ 0.023}  &  &  &  \\ \cline{2-4}
\multicolumn{1}{|c|}{}                        & \multicolumn{1}{l|}{1-GPA}    & \multicolumn{1}{l|}{0.918763 $\pm$ 0.002} & \multicolumn{1}{l|}{0.925992 $\pm$ 0.008}  &  &  &  \\ \cline{2-4}
\multicolumn{1}{|c|}{}                        & \multicolumn{1}{l|}{nDCG@500} & \multicolumn{1}{l|}{0.671236 $\pm$ 0.005} & \multicolumn{1}{l|}{0.652672 $\pm$ 0.014}  &  &  &  \\ \cline{1-4}
\multicolumn{1}{l}{}                          &                               &                                           &                                            &  &  &  \\
\multicolumn{1}{l}{}                          &                               &                                           &                                            &  &  &  \\
\end{tabular}
\caption{Results for the experimentation performed on the COMPAS and Bank datasets. For all the metrics, higher is better. We observe that the performance of the Fair DirectRanker (Fair DR) by \cite{cerrato2020constraining} is not impacted meaningfully when constrained for interpretability.}
\end{table}

\subsection{The COMPAS dataset}

We focus our discussion on the COMPAS dataset, one of the most popular datasets in fair classification and ranking. This dataset has been published by ProPublica (\cite{machine_bias}) after a long-term evaluation of the COMPAS tool, short for ``Correctional Offender Management Profiling for Alternative Sanctions''. This tool is made available to US judges, who may employ it when deciding to allow an individual to be released on parole. The rationale here is that an individual that is evaluated as ``low risk'' could be allowed to pay bail and avoid incarceration while waiting for trial. The opposite follows for individuals that are deemed ``high risk''. ProPublica evaluated that the tool is biased against black people in the sense that it mis-assigns high risk scores to black people on a higher rate than to white people. We employ, as widely done in the literature, the 10 COMPAS classes as relevance classes for our ranking algorithm. On top of evaluating relevance (via nDCG) and fairness (via rND), we analyze the correction vectors focusing on the ``priors\_count''. This feature represents the number of previous crimes committed by an individual, and investigating how a fair model changes this feature over the two different groups available (white and black people) can provide insights into the model's reasoning. We provide the average correction value for our best model in Table~2. In the table, we observe that making the two groups more similar translates into \emph{disparate corrections} which impact black people more than white. Here, one could make the argument that changing the attributes of individuals is equivalent to rewriting their personal history, and could be seen as unlawful. This objection merits attention, and needs to be investigated further. At this time, we would posit that this issue is common to all fair representation learning algorithms, with the difference that the correction is computed by projecting individuals' data into non-readable latent dimensions. The benefit of our framework is that it is possible to investigate this transformation, and possibly refuse the decisions if they are seen as problematic.

\begin{table}[]
\label{tab:interpretable-vects}
\centering
\begin{tabular}{l|l|l|l|}
\cline{2-4}
                                               & Black             & White             & Avg. Difference   \\ \hline
\multicolumn{1}{|l|}{priors\_count, original}  & 2.406494          & 1.578894          & 0.8276            \\ \hline
\multicolumn{1}{|l|}{priors\_count, corrected} & 2.211587 (-8.1\%) & 1.486147 (-5.8\%) & 0.72544 (-12.4\%) \\ \hline
\end{tabular}
\caption{Average values for the \emph{priors\_count} feature in the COMPAS dataset over the two ethnicity groups. We observe \emph{disparate corrections}, i.e. black individuals receive a stronger negative correction.}
\end{table}

\section{Conclusions and Future Work}

In this paper, we presented a new framework for interpretable fair representation learning, which computes correction vectors. Our experimentation shows that losses in performance and fairness metrics are negligible when constraining a state-of-the-art fair ranker for interpretability. One point that needs further investigation is whether learning corrections is a desirable property. While we argue that all representation algorithms learn corrections -- usually by projecting into a human-unreadable latent space -- we are currently investigating this matter, collaborating with experts in IT law and equality rights.

\end{document}